\documentclass{article} 
\PassOptionsToPackage{dvipsnames}{xcolor}
\usepackage{iclr2025_conference,times}

\usepackage{amsmath,amsfonts,bm}

\def\eqref#1{equation~\ref{#1}}

\def\1{\bm{1}}

\def\vx{{\bm{x}}}

\DeclareMathAlphabet{\mathsfit}{\encodingdefault}{\sfdefault}{m}{sl}
\SetMathAlphabet{\mathsfit}{bold}{\encodingdefault}{\sfdefault}{bx}{n}

\usepackage{hyperref}
\usepackage[capitalise]{cleveref}
\usepackage{url}
\usepackage{amssymb}
\usepackage{enumitem}
\usepackage[ruled,noend]{algorithm2e}
\usepackage{booktabs}
\usepackage{adjustbox}
\usepackage{wrapfig,lipsum}
\usepackage{floatrow}
\usepackage{scalerel}
\usepackage{subfigure}

\usepackage{graphicx}

\makeatletter
\def\thanks#1{\protected@xdef\@thanks{\@thanks
        \protect\footnotetext{#1}}}
\makeatother

\hypersetup{
colorlinks=true,linkcolor=Salmon,citecolor=teal,urlcolor=orange
}

\title{ Scalable Expectation Estimation\\with Subtractive Mixture Models}

\author{$\text{Lena Zellinger}^{\boldsymbol{\Delta}}, \text{Nicola Branchini}^{\boldsymbol{\Delta}}$, V\'ictor Elvira, Antonio Vergari  \thanks{ $\boldsymbol{\Delta}$: Shared first authorship.} \\
School of Informatics \& School of Mathematics\\
University of Edinburgh
}

\newcommand{\method}{\Delta\text{Ex}}

\newtheorem{proposition}{Proposition}

\begin{document}

\maketitle

\begin{abstract}
Many Monte Carlo (MC) and importance sampling (IS) methods use mixture models (MMs) for their simplicity and ability to capture multimodal distributions. 
Recently, \textit{subtractive mixture models} (SMMs), i.e. MMs with negative coefficients, have shown greater expressiveness and success in generative modeling.
However, their negative parameters complicate sampling, requiring costly auto-regressive techniques or accept-reject algorithms that do not scale in high dimensions.
In this work, we use the difference representation of SMMs to construct
an unbiased IS estimator ($\Delta\text{Ex}$) that removes the need to sample from the SMM, enabling high-dimensional expectation estimation with SMMs.
In our experiments, we show  that $\Delta\text{Ex}$ can achieve comparable estimation quality to auto-regressive sampling while being considerably faster in MC estimation. Moreover, we conduct initial experiments with $\Delta\text{Ex}$ using hand-crafted proposals, gaining first insights into how to construct safe proposals for $\Delta\text{Ex}$. %
\end{abstract}

\section{Introduction}
\label{sec:introduction}

Many tasks in probabilistic machine learning (ML) and statistics amount to approximating expectations under complex probability distributions, including computing average treatment effects \citep{hirano2003efficient,khan2023adaptive}, several fairness metrics \citep{zhang2021assessing}, and predictions in Bayesian inference \citep{vehtari2017practical}. 
Such intractable expectations are commonly estimated via \emph{importance sampling (IS)}, which is a generalization of standard Monte Carlo (MC) integration.
IS allows to approximate expectations under a distribution $p$ by sampling from a so-called \emph{proposal distribution} $q$ while preserving unbiasedness.
This can be particularly beneficial when sampling from $p$ directly is costly or results in high variance.
To achieve low-variance estimators with IS, the chosen family of proposal distributions should be expressive enough to closely model the target of integration while also supporting efficient sampling.

Mixture models (MMs) are a natural choice of proposals in IS for their simplicity and ability to represent multimodal distributions \citep{owen2000safe,bugallo2017adaptive}.
Classical MMs are however fundamentally restricted to \textit{add} probability densities.
This implies that, in many scenarios, they require a large number of components to accurately represent the target distribution.
Recently, \textit{subtractive mixture models} (SMMs), which allow for negative mixture weights, have gained attention in probabilistic ML \citep{marteauferey2020nonparametric,rudi2021-psd,loconte2024subtractive,cai2024eigenvi}. Due to their ability to subtract densities, SMMs can represent complex distributions while provably requiring exponentially less components than classical additive MMs \citep{loconte2024subtractive,loconte2024sos}. However, sampling techniques for SMMs, such as auto-regressive inverse transform sampling (ARITS) \citep{loconte2024subtractive,cai2024eigenvi} 
and accept-reject methods \citep{robert2025simulating},
can be computationally expensive and hence unsuitable for high dimensional expectation estimation.

In this preliminary work, we motivate the use of SMMs for IS, highlighting their expressiveness and theoretical connection to optimal IS proposals. Moreover, we discuss how to make expectation estimation with SMMs feasible in practice. 
To this end, we propose $\Delta\text{Ex}$ - an unbiased IS estimator which avoids costly sampling from the SMM by writing an SMM  as a \emph{difference of additive MMs} \citep{bignami1971note,robert2025simulating}.
We show empirically that $\Delta\text{Ex}$ can achieve similar estimation quality to costly ARITS sampling and discuss how to further reduce its variance. Moreover, we test $\Delta\text{Ex}$ with hand-crafted proposals revealing that a low Kullback-Leibler (KL) divergence between the proposal and target of integration does not necessarily result in a low-variance estimator. %

\textbf{Contributions.}
\textbf{(i)} We study for the first time, to the best of our knowledge, the use of SMMs to estimate expectations with IS. 
\textbf{(ii)} We propose and analyze a new IS estimator, $\Delta\text{Ex}$, that avoids costly sampling from the proposal SMM, thereby lowering the computational complexity by a factor of $d$, the ambient dimension, and discuss how to further reduce its variance. 
\textbf{(iii)} We conduct experiments with $\Delta\text{Ex}$ in a standard MC setting, demonstrating its computational efficiency compared to ARITS. We further test $\Delta\text{Ex}$ with various synthetic proposals and propose initial directions for obtaining low-variance estimators with $\Delta\text{Ex}$. 

\section{Subtractive Mixture Models}
\label{sec:squared-smm}

The recipe for classical additive MMs is simple: combine valid probability density functions (PDFs) in a convex combination \citep{mclachlan2019finite}.
Popular choices of base distributions for a mixture component are exponential families, or more expressive probabilistic models 
such as normalizing flows \citep{pires2020variational}.
As their formulation only allows to \textit{add} PDFs, to recover one target PDF of interest they can require an exponential number of components.
\begin{figure}
    \centering
    \raisebox{20pt}{\scalebox{6}{{(}}}\includegraphics[height=.15\textwidth, page=1]{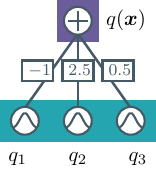}
    \raisebox{20pt}{\scalebox{6}{{)}}\raisebox{30pt}{\scalebox{1.5}{$^2$}}}
    \raisebox{30pt}{\scalebox{3}{=}}
    \includegraphics[height=.15\textwidth, page=2]{figures/smm.pdf}
    \raisebox{30pt}{\scalebox{3}{=}}
    \includegraphics[height=.15\textwidth, page=3]{figures/smm.pdf}
    \raisebox{30pt}{\scalebox{3}{$-$}}
    \includegraphics[height=.15\textwidth, page=4]{figures/smm.pdf}

    \caption{
    \textbf{A squared mixture can be split into its positive and negative parts} as illustrated via its representation as a computational graph, also called circuit \citep{choi2020probabilistic,loconte2024relationshiptensorfactorizationscircuits}.
    }
    \label{fig:smms}
\end{figure}

Intuitively, SMMs increase the \textit{expressive efficiency} \citep{choi2020probabilistic} of classical additive MMs by dropping the convex constraint over the mixture coefficients.
An SMM over $\vx$ is thus given by
\begin{equation}
    \tag{SMM}
    \label{eq:smm}
    q_{\text{SMM}}(\boldsymbol{x}) = Z_q^{-1} \cdot \sum\nolimits_{k=1}^{K} \alpha_{k} q_{k}(\boldsymbol{x}), ~ \text{where}
    ~ ~ \text{and} ~~ \alpha_{k}  \in \mathbb{R}, \ \text{for} ~ k=1,\dots, K,
\end{equation}
where $Z_q=\sum\nolimits_{k=1}^{K} \alpha_{k}\int q_k(\boldsymbol{x})d\boldsymbol{x}$ is the normalizing constant and the mixture coefficients $\alpha_k$ are allowed to be negative.
A challenge when \textit{learning} SMMs is constraining $q(\boldsymbol{x}) \geq 0$ to retain a valid PDF. 
While it is possible to derive closed-form constraints for simple parametric $q_k$, such as Gaussians, Gammas and Weibulls \citep{jiang1999weibull,zhang2005finite,rabusseau2014learning}, this is non-trivial in general.
To this end, 
\cite{loconte2024subtractive} developed squared SMMs, ensuring the non-negativity of $q$ by squaring \cref{eq:smm}, i.e.,
\begin{equation}\label{eq:ssmm}
    q_{\text{SMM}^2}(\boldsymbol{x}) 
    = Z_{q}^{-1} \cdot \left(\sum\nolimits_{k=1}^{K}\alpha_{k} q_{k}(\boldsymbol{x})\right)^2
    = Z_{q}^{-1} \cdot \sum\nolimits_{k=1}^{K}\sum\nolimits_{k^\prime=1}^{K^\prime} \alpha_{k}\alpha_{k^\prime} q_{k}(\boldsymbol{x}) q_{k^\prime}(\boldsymbol{x})   ,
\end{equation}
where $Z_{q} =  \int \sum_{k=1}^{K}\sum_{k^\prime=1}^{K^\prime} \alpha_{k}\alpha_{k^\prime} q_{k}(\boldsymbol{x}) q_{k^\prime}(\boldsymbol{x}) d \vx $. 
\cref{fig:smms} shows the computational graph of a squared SMM.
Note that \cref{eq:ssmm} is still an SMM, as some $\alpha_{k}\alpha_{k^\prime}$ can be $< 0$, but $q$ is guaranteed to be $\geq 0$.\footnote{While a squared SMM has up to ${K+1}\choose{2}$ unique components due to the pairwise products, the number of learnable parameters remains the same as for its non-squared counterpart \citep{loconte2024subtractive}.} 
To exactly compute $Z_q$, the products $q_{k}q_{k^\prime}$ need to be integrable, a condition verified for exponential families and other functions such as polynomials \citep{loconte2024subtractive}.
Squared SMMs can be further combined into a sum of squared (SOS) SMMs, which can be proven to be more expressive efficient than both additive MMs and squared SMMs \citep{loconte2024sos}.
Analogously to squared SMMs, SOS SMMs can be easily rewritten in the form of \cref{eq:smm}. Therefore, from here on we will assume a valid SMM to be in such a form.

\section{Importance Sampling with SMMs}
\label{sec:sampling}

IS is based on the idea that we can estimate expectations over $p$ by drawing i.i.d. samples from a proposal PDF $q$ of choice. 
The unnormalized IS (UIS) estimator of the integral $I = \int f(\vx) p(\vx) d \vx$ for a function $f: \mathcal{X} \rightarrow \mathbb{R}$ such that $\int_{\mathcal{X}} |f(\vx)| p(\vx) < \infty$ is given by 
\begin{equation}\label{eq:uis}
    \widehat{I}_{\text{UIS}} = \frac{1}{S} \sum\nolimits_{s=1}^{S} w^{(s)} f(\boldsymbol{x}^{(s)}) , \quad\text{where}\ \ w^{(s)} = \frac{p(\vx^{(s)})}{q(\vx^{(s)})}, ~~ \text{and}\quad\boldsymbol{x}^{(s)} \sim q(\vx),
\end{equation}
where $p \ll q$. When $q(\vx) = p(\vx)$, $\widehat{I}_{\text{UIS}}$ reduces to simple MC. 
The optimal (variance-minimizing) proposal for \cref{eq:uis} is $q^{\bigstar}_{\text{UIS}}(\vx) = Z_{q^{\bigstar}}^{-1} \cdot |f(\vx)| p(\vx)$ \citep{robert1999monte,mcbook}, which can be very different from $p$. 
The interest in using SMMs for importance sampling is natural not only due to their expressiveness, but also because the IS optimal proposal often explicitly takes the form of an SMM. %
For instance, this is the case when one is interested in a difference of expectations, i.e., quantities in the form $\mathbb{E}_{x \sim p_0(\boldsymbol{x})}[f_{0}(\boldsymbol{x})] - \mathbb{E}_{x \sim p_1(\boldsymbol{x})}[f_{1}(\boldsymbol{x})]$, such as the average treatment effect in causal inference \citep{hirano2003efficient}.
The optimal UIS proposal in this case is easily shown to be proportional to $|p_{0}(\vx) f_{0}(\vx) - p_{1}(\vx) f_{1}(\vx)|$, and hence takes the form of an SMM.

\textbf{SMMs and the SNIS estimator.} If $p$ is only known up to a normalizing constant one can resort to the self-normalized IS estimator (SNIS). Let $p(\vx) = Z_{p}^{-1} \cdot \widetilde{p}(\vx)$, then
\begin{equation}\label{eq:snis}
    \widehat{I}_{\text{SNIS}} = \sum\nolimits_{s=1}^{S} \overline{w}^{(s)} f(\boldsymbol{x}^{(s)}) , ~~ \overline{w}^{(s)} = {\widetilde{w}^{(s)}}\Big/\left({\sum\nolimits_{i=1}^{S}\widetilde{w}^{(i)}}\right) , ~ \widetilde{w}^{(s)} = \frac{\widetilde{p}(\vx^{(s)})}{q(\vx^{(s)})} .
\end{equation}
For SNIS the proposal $q \in \mathcal{P}(\mathcal{X})$ that minimizes the asymptotic variance is known to be
\begin{equation}\label{eq:optsnisproposal}
    q^{\bigstar}_{\text{SNIS}}(\vx) = Z_{q^{\bigstar}_{\text{SNIS}}}^{-1} \cdot |p(\vx) f(\vx) - I \cdot p(\vx)| ,
\end{equation}
where commonly $\mathcal{P}(\mathcal{X})$ is the set of square integrable PDFs, i.e., $\int q(\vx)^2 d \vx < \infty$ \citep{geweke1989bayesian,rainforth2020target,branchini2024generalizing}.
Interestingly, \cref{eq:optsnisproposal} takes the form of an SMM, where the first component is the optimal UIS proposal, while the second is just $p(\vx)$ scaled by $I$.

\subsection{How to sample an SMM?} 
Although using SMMs as proposals is promising, sampling from them is known to be more challenging than sampling from MMs.
This is because, for the latter, one can use the latent variable interpretation of MMs \citep{peharz2016latent} and perform ancestral mixture sampling \citep{bishop2006pattern} (see also \cref{app:sampling_techniques}), while for SMMs, the introduction of negative coefficients breaks such a latent variable interpretation.
Nevertheless, it is possible to sample from SMMs, e.g., by performing auto-regressive inverse transform sampling (ARITS) \citep{loconte2024subtractive,cai2024eigenvi} or using accept-reject algorithms \citep{robert2025simulating}. 

Unfortunately, neither of these approaches scales as gracefully as ancestral sampling for additive MMs. 
For instance, ARITS entails fixing a variable ordering and sampling \emph{iteratively},
$x_1 \sim q_{\text{SMM}}(x_1)$ and $x_i \sim q_{\text{SMM}}(x_i|x_{<i})$ for $i \in \{2,...,d\}$, where each sampling step requires numerically inverting the corresponding conditional cumulative distribution function (CDF).
As a result, ARITS has a cost that is $d$-times slower than classical ancestral sampling, assuming that computing the univariate marginals and inverting the CDFs costs equivalently to sampling one component in a MM. 
See Algorithm \ref{alg:ARITS} in \cref{sec:appendix_ARITS} for the pseudocode of ARITS. 
In the following, we show how to make the use of SMMs as proposals in a  computationally feasible way:
we devise an estimator that recovers the cost of ancestral sampling while providing unbiased estimates.  

\section{Avoiding sampling from SMMs: the $\method$ estimator}\label{subsec:estimator} 
Following \citet{robert2025simulating}, we rewrite an SMM as the \textbf{\emph{difference of two {additive} MMs}}, i.e.,
\begin{equation}
\label{eq:diffrep}
    q(\vx) = \frac{1}{Z_q}\Big( Z_{+} \cdot q_{+}(\vx) - Z_{-} \cdot q_{-}(\vx)\Big) ,
\end{equation}
where $q_{+}$ and $q_{-}$ are composed of the positively and negatively weighted components of $q$ respectively, $Z_{+}, Z_{-}$ are their normalizing constants, and $Z_q = Z_+ -Z_-$ is the normalizing constant of $q$.
Such a difference representation has been known in previous works \citep{bignami1971note,rabusseau2014learning,robert2025simulating}, but to the best of our knowledge they did not evaluate the quality of practical MC estimators nor considered extending them to IS. 
Converting a sum of squared SMMs (\cref{eq:ssmm}) into \cref{eq:diffrep} is always possible and can be done in time linear in the number of components (after squaring).
\cref{fig:smms} shows an example.
Such a  representation allows us \textit{to split an expectation over the SMM into a difference of expectations, each of which is an expectation w.r.t. an additive MM}.
Therefore, $I$ can be rewritten as 
\begin{equation}\label{eq:diffrep_UIS}
I = \int f(\vx) p(\vx) d \vx = \mathbb{E}_{q} \left [ f(\vx)w(\vx)  \right ] = \frac{Z_{+}}{Z_q} \mathbb{E}_{q_{+}} \left [ f(\vx)w(\vx) \right ] - \frac{Z_{-}}{Z_q} \mathbb{E}_{q_{-}} \left [ f(\vx)w(\vx)\right ],
\end{equation}
where $w(\vx)={p(\vx)}/{q(\vx)}$. This leads to our proposed \textbf{\textit{difference of expectations estimator}} ($\method$):
\begin{equation}\label{eq:our_estimator}
\widehat{I}_{\method} = \frac{Z_+}{Z_q}  \frac{1}{S_{+}} \sum\nolimits_{s=1}^{S_{+}}  f(\vx_{+}^{(s)}) w(\vx_{+}^{(s)}) - \frac{Z_-}{Z_q} \frac{1}{S_{-}} \sum\nolimits_{s=1}^{S_{-}} f(\vx_{-}^{(s)})w(\vx_{-}^{(s)}), 
\begin{array}{l}
\vx_{+}^{(s)}\sim q_{+}(\vx_{+}) \\[2pt]
\vx_{-}^{(s)} \sim q_{-}(\vx_{-})
\end{array},
\end{equation}
where $S = S_{+} + S_{-}.$ Note that, interestingly, while we sample from $q_{+}$ and $ q_{-}$, the weights have the full SMM $q$ in the denominator.
With $\method$, we can sample both $q_{+}$ and $q_{-}$ via classical ancestral sampling with no need to approximate inverse CDFs and iterate $d$ times.
Proposition \ref{proposition:properties} states the core properties of our proposed estimator. \cref{fig:sample_comparison} in \cref{app:sample_comparison} illustrates samples used for expectation estimation with $\Delta\text{Ex}$ as well as samples obtained by ARITS.
\begin{proposition}[Properties of $\method$]\label{proposition:properties} See \cref{app:proofs} for proofs.
\begin{enumerate}[leftmargin=2pt]
\setlength{\itemindent}{12pt}
\setlength\itemsep{0.5pt}
    \item \textbf{Unbiasedness and strong consistency:} the $\method$ estimator is unbiased, i.e., $\mathbb{E}_{\substack{\vx_+ \sim q_+\\\vx_- \sim q_-}}[\widehat{I}_{\method}] = I$, and it is strongly consistent, $\mathbb{P}_{\substack{\vx_+ \sim q_+\\\vx_- \sim q_-}}\left ( \lim_{\substack{S_+ \rightarrow +\infty\\S_- \rightarrow +\infty}} \widehat{I}_{\method} = I \right ) = 1$.
    \item \textbf{Variance expression:} for simplicity, let $I = \int \widetilde{p}(\vx) d \vx = Z_{p}$; an analogue expression for generic $f(\vx)$ is in \cref{app:proofs}. Let $\widetilde{w}(\vx)={\widetilde{p}(\vx)}/{q(\vx)}$.
The variance of $\widehat{I}_{\method}$ is given by
\begin{align}\label{eq:var}
      \resizebox{0.9\hsize}{!}{$ \frac{Z_{+}^{2}}{Z_{q}^{2}}\frac{1}{S_{+}} \left ( \mathbb{E}_{\vx \sim q_{+}}\left [ \left (\widetilde{w}(\vx) \right )^2 \right ] - \left(\mathbb{E}_{\vx \sim q_{+}}\left [ \widetilde{w}(\vx) \right ]\right)^2 \right ) + \frac{Z_{-}^{2}}{Z_{q}^{2}} \frac{1}{S_{-}} \left ( \mathbb{E}_{\vx \sim q_{-}}\left [ \left (\widetilde{w}(\vx) \right )^2 \right ] - \left(\mathbb{E}_{\vx \sim q_{-}}\left [ \widetilde{w}(\vx) \right ]\right)^2 \right ) . $} \nonumber 
   \end{align} 
   Necessary and sufficient conditions for finite variance are: \textbf{(i)} $q(\vx) \neq 0$ almost-everywhere in the support of $q_{+}$ and $q_{-}$; \textbf{(ii)} $\int\left( \widetilde{w}(\vx) \right)^2 q_{+}(\vx) d \vx<\infty$ and $\int\left( \widetilde{w}(\vx) \right)^2 q_{-}(\vx) d \vx<\infty$.
\end{enumerate}
\end{proposition}
Note that for $\method$, the variance is more complex than for the standard UIS estimator. First, more terms appear as $\method$ is a difference of two estimators that, individually, are \emph{biased} for $I$ - yet, unbiased when combined. Second, while the importance weight $\widetilde{w}(\vx) = \widetilde{p}(\vx)/q(\vx)$ evaluates the full SMM in the denominator, the expectations are under $q_{+}$ and $q_{-}$ instead of $q$. This complicates interpreting the variance as a known divergence between $q$ and $p$. However, it shares a basic property with divergences $\mathrm{D}(p;q)$: if $p = q$, it is zero. Detailed intuitions about which properties a good proposal $q$ should satisfy will require further work.

\textbf{Variance reduction.} Since $\method$ splits the SMM into two additive mixtures, it is straightforward to apply additional variance reduction techniques. For instance, we can use \textit{stratified sampling} \citep{mcbook} to draw a deterministic number of samples from each component of $q_+$ and $q_-$ respectively. Stratified sampling is known to reduce the variance of the standard UIS estimator compared to using ancestral sampling \citep{hesterberg1995weighted}.\footnote{This is also known as \emph{deterministic mixture} sampling in the multiple IS (MIS) literature \citep{owen2000safe,elvira2019generalized} } 
We choose stratified sampling as a first, easy to implement variance reduction technique for our experiments (\cref{sec:experiments}), and refer to the stratified variant of our method as $\Delta\text{ExS}$, while $\Delta\text{ExA}$ uses ancestral sampling. A proof showing that $\text{Var}[\widehat{I}_{\Delta\text{ExS}}] \leq \text{Var}[\widehat{I}_{\Delta\text{ExA}}]$ can be found in \cref{app:sampling_techniques}. However, we did not find strong evidence for this variance reduction in our experiments (see \cref{app:variance_comparison}).

\textbf{Sampling budget.} When computing $\widehat{I}_{\Delta\text{Ex}}$ for a given budget $S$, we propose to fix the sample size for the two mixtures as $S_+ = \big\lfloor\frac{Z_+}{Z_+ + Z_-} S\big\rfloor$ and $S_- = \big\lfloor\frac{Z_-}{Z_+ + Z_-} S\big\rfloor$. Intuitively, we sample from the mixtures in proportion to their relative contribution to the estimator. This is the sample allocation we use in \cref{sec:experiments}. For an empirical comparison with equal sample allocation, see \cref{app:sampling_budget}. 

\subsection{A Safer $\method$}\label{sec:safe_dex}

As just discussed, our $\method$ method provides more scalable computation than ARITS-based estimators while maintaining several nice statistical properties.
At the same time, it can pose additional challenges that do not appear when sampling from the SMM directly.
To see why, consider a scenario where some of the samples obtained by $\Delta{\text{Ex}}$ evaluate to very close, non-zero values for \textit{both} $\frac{Z_+}{Z_q} q_{+}$ \textit{and} $\frac{Z_-}{Z_q} q_{-}$. 
For such samples, the resulting density $q$ would be close to zero, potentially leading to exploding importance weights, and with the crucial consequence of resulting in high-variance estimates. 
\cref{fig:synthetic_proposals} shows a concrete example: The second target density and its proposals have deep, low-density valleys and result in high variance for $\Delta\text{Ex}$.
Note that this potential issue is dependent on the chosen proposal but it differs from the usual reason for high variance
that can appear in IS estimators \citep{delyon2021safe}. Typically in IS, one can have a large weight (hence, large variance) because the proposal has lighter tails than the target, so that for some samples, $q(\vx) $ is much smaller than $p(\vx)$. Our problem is specific to $\Delta\text{Ex}$ since it samples where $q_+$ and $q_-$ individually have high density, even if the overall $q$ there is close to 0. ARITS, on the other hand, is unlikely to sample in areas where $q$ is close to 0 (see \cref{fig:sample_comparison}). %

To improve the variance of our estimator in such scenarios, we take inspiration from safe adaptive IS (SAIS) \citep{delyon2021safe,korba2022adaptive,bianchi2024stochastic} and include a ``safe'' mixture component in our proposal. Given a SMM proposal $q_\text{SMM}$, we mix it with the safe component $q_{\text{safe}}$ in a convex combination, resulting in $q(\vx) = (1-\alpha)q_{\text{SMM}}(\vx) + \alpha q_{\text{safe}}(\vx)$, where $\alpha$ is a hyperparameter and the sampling budget can be allocated as $S_{q_\text{SMM}} = \lfloor(1-\alpha)S\rfloor$ and $S_{q_\text{safe}} = \lfloor\alpha  S\rfloor$.
Differently from the SAIS literature, the ``safe'' component here does not necessarily need to be a heavy-tailed density, but rather a flat one that covers the low-density valleys of the SMM (see \cref{fig:synthetic_proposals}).
In the next section, we show how this simple modification yields much better estimates for target densities and proposals that would otherwise be problematic with $\method$.

\section{Experiments}
In this Section, we conduct preliminary experiments with $\method$ in synthetic examples to investigate the following research questions: (\textbf{RQ1}): \emph{``How does $\method$ compare to ARITS in terms of error and runtime?''}, 
and (\textbf{RQ2}): \emph{``What is the intuition for a good proposal for $\method$ ?''}.
By doing so, we evaluate the potential of $\method$ for large-scale expectation estimation in a controlled setting and motivate further investigation on diverse problems. 
\label{sec:experiments}

\subsection{RQ1: Runtime and Estimation Quality of $\Delta\text{Ex}$ and ARITS}\label{sec:runtime_comparison}
We compare ARITS and $\Delta\text{Ex}{\text{S}}$ in terms of runtime and estimation quality
in a standard MC setting. We estimate $I = \mathbb{E}_p[f(\vx)]$, where $p$ is a squared SMM with Gaussian inputs, and $f$ is a standard GMM defined over the same set of variables as $p$, so we can compute the ground-truth value of $I$ exactly. We vary the number of variables $d$ (dimension) in $\{16, 32, 64\}$. Moreover, we let the number of components in $p$ range in $\{2, 4, 6\}$ before squaring\footnote{Note that this results in up to 3, 10 and 21 unique components after squaring respectively.}, while keeping it fixed to $100$ for $f$ to encourage non-vanishing values of $I$ in higher dimensions. The budget for ARITS estimators is fixed to $10000$ samples while $\Delta\text{ExS}$ estimators are computed for $S \in \{10000, 100000, 300000\}$. The results are averaged over $30$ different initializations of $f$ and $p$. We measure the estimation error as $\log(| \widehat{I} -  I|)$ and the runtime is reported in seconds.
Further details on the initialization of $p$ and $f$, the used hardware, and the ARITS implementation can be found in \cref{sec:experiments_appendix}.
\begin{table}
\resizebox{\textwidth}{!}{
\begin{tabular}{lrrrrrrrr}
\toprule
&&&\multicolumn{6}{c}{\textbf{Number of components $(\boldsymbol{K})$}}\\
\cmidrule(lr){4-9}
\multicolumn{3}{c}{} & \multicolumn{2}{c}{$\boldsymbol{2}$} & \multicolumn{2}{c}{$\boldsymbol{4}$} & \multicolumn{2}{c}{$\boldsymbol{6}$} \\
\cmidrule(lr){4-5} \cmidrule(lr){6-7} \cmidrule(lr){8-9}
\textbf{Method} &  \boldsymbol{$d$} & \boldsymbol{$S$} & $\log(| \widehat{I} -  I|)$ & Time (s) & $\log(| \widehat{I} -  I|)$ & Time (s) & $\log(| \widehat{I} -  I|)$ & Time (s) \\
\midrule
$\Delta\text{ExS}$ & 16 & 10000 & -18.650 $\pm$ 1.291 & 0.053 $\pm$ 0.072 & -18.240 $\pm$ 1.314 & 0.102 $\pm$ 0.007 & -17.942 $\pm$ 1.199 & 0.219 $\pm$ 0.026 \\
$\Delta\text{ExS}$ & 16 & 100000 & -19.722 $\pm$ 1.041 & 0.238 $\pm$ 0.004 & -19.231 $\pm$ 1.228 & 0.659 $\pm$ 0.039 & -19.009 $\pm$ 1.379 & 1.368 $\pm$ 0.087 \\
$\Delta\text{ExS}$ & 16 & 300000 & -19.937 $\pm$ 0.797 & 0.668 $\pm$ 0.013 & -19.743 $\pm$ 1.070 & 1.862 $\pm$ 0.112 & -19.491 $\pm$ 0.931 & 3.851 $\pm$ 0.226 \\
ARITS & 16 & 10000 & -19.111 $\pm$ 1.103 & 7.579 $\pm$ 0.120 & -19.299 $\pm$ 1.611 & 7.588 $\pm$ 0.037 & -18.739 $\pm$ 1.024 & 7.869 $\pm$ 0.034 \\ \midrule
$\Delta\text{ExS}$ & 32 & 10000 & -48.024 $\pm$ 1.344 & 0.042 $\pm$ 0.003 & -47.231 $\pm$ 1.059 & 0.097 $\pm$ 0.008 & -47.273 $\pm$ 1.015 & 0.185 $\pm$ 0.017 \\
$\Delta\text{ExS}$ & 32 & 100000 & -48.395 $\pm$ 1.068 & 0.268 $\pm$ 0.009 & -48.092 $\pm$ 0.771 & 0.652 $\pm$ 0.055 & -48.426 $\pm$ 1.734 & 1.313 $\pm$ 0.091 \\
$\Delta\text{ExS}$ & 32 & 300000 & -48.729 $\pm$ 0.835 & 0.762 $\pm$ 0.030 & -48.671 $\pm$ 1.758 & 1.869 $\pm$ 0.162 & -48.791 $\pm$ 1.007 & 3.756 $\pm$ 0.259 \\
ARITS & 32 & 10000 & -47.897 $\pm$ 1.165 & 15.460 $\pm$ 0.116 & -47.349 $\pm$ 0.839 & 15.730 $\pm$ 0.058 & -47.300 $\pm$ 0.978 & 17.560 $\pm$ 0.060 \\ \midrule
$\Delta\text{ExS}$ & 64 & 10000 & -108.129 $\pm$ 1.280 & 0.059 $\pm$ 0.061 & -107.338 $\pm$ 0.854 & 0.085 $\pm$ 0.008 & -107.444 $\pm$ 0.827 & 0.137 $\pm$ 0.022 \\
$\Delta\text{ExS}$ & 64 & 100000 & -108.000 $\pm$ 1.066 & 0.331 $\pm$ 0.017 & -107.431 $\pm$ 0.692 & 0.621 $\pm$ 0.070 & -107.672 $\pm$ 0.782 & 0.980 $\pm$ 0.194 \\
$\Delta\text{ExS}$ & 64 & 300000 & -108.354 $\pm$ 1.436 & 0.961 $\pm$ 0.049 & -107.569 $\pm$ 1.215 & 1.801 $\pm$ 0.207 & -108.052 $\pm$ 1.372 & 2.845 $\pm$ 0.576 \\
ARITS & 64 & 10000 & -107.898 $\pm$ 1.129 & 31.036 $\pm$ 0.115 & -107.330 $\pm$ 0.929 & 34.318 $\pm$ 0.142 & -107.374 $\pm$ 1.138 & 52.490 $\pm$ 0.095 \\
\bottomrule
\end{tabular}
}
\caption{\textbf{$\Delta\text{ExS}$ is consistently faster than ARITS for expectation estimation while achieving comparable estimation quality.} Results for MC comparing our method ($\Delta\text{ExS}$) with ARITS for a varying number of features ($d$) and components ($K$) as well as different sampling budgets ($S$) for $\Delta\text{ExS}$. The error is given as $\log(| \widehat{I} -  I|)$, hence lower is better, and time is in seconds. Results are averaged over 30 initializations of $p$ and $f$ (mean $\pm$ stddev).}\label{tab:exp_1}
\vspace*{-\baselineskip}
\end{table}

Table \ref{tab:exp_1} shows that $\Delta\text{ExS}$ is consistently faster than ARITS, even when using a much larger sampling budget. 
At the same time, the estimation quality of $\Delta\text{ExS}$ is generally comparable to ARITS when a sufficiently large sample size is used. These results highlight the potential of $\Delta\text{ExS}$ for high-dimensional expectation estimation where ARITS is computationally infeasible.
\subsection{RQ2: $\Delta\text{Ex}$ for Normalizing Constant Estimation}
To address our second research question, we test $\Delta\text{ExS}$ for normalizing constant estimation using hand-crafted proposals. In particular, given an unnormalized target distribution $\widetilde{p}$, we aim to approximate $I =\int \widetilde{p}(\vx)d\vx=\mathbb{E}_{q}[\widetilde{p}(\vx)/q(\vx)]$.
Note that the task of normalizing constant estimation requires the use of IS, as opposed standard MC, since the problem is not an expectation under $p$.

Remember that the optimal proposal in standard UIS is known to be $q^{\bigstar}_{\text{UIS}}(\vx) = Z_{q^{\bigstar}}^{-1} \cdot |f(\vx)| p(\vx)$ \citep{robert1999monte,mcbook}, which reduces to $q^{\bigstar}_{\text{UIS}}(\vx) = p(\vx)$ for normalizing constant estimation.
We therefore create proposals by slightly noising the standard deviations of the Gaussian components of $p$, which should result in proposals that are fairly close to the optimal UIS proposal.
In particular, we set $\sigma_q := \sigma_p \cdot \exp(\epsilon \cdot Z)$, where $Z \sim \mathcal{N}(0, 1)$ and $\epsilon \geq 0$, for each input standard deviation $\sigma_p$. We apply these perturbations before squaring the proposal.
Moreover, we want to understand the the potential benefit of including a safe mixture component, as described in \cref{sec:safe_dex}. We heuristically choose $q_{\text{safe}}$ as a zero-mean, zero-covariance multivariate Gaussian with a large standard deviation compared to the other components of the proposal, namely $\sigma=3$, and set $\alpha=0.001$.

\begin{figure}[!t]
\begin{center}
\setlength{\subfigcapskip}{0pt}  %
\setlength{\subfigtopskip}{0pt}  %
\setlength{\subfigbottomskip}{0pt}  %
\setlength{\tabcolsep}{2pt}

\begin{tabular}{rrrrr}
& \multicolumn{2}{c}{\small$q_{\text{SMM}}(\vx)$} & \multicolumn{2}{c}{\small$0.999\cdot q_{\text{SMM}}(\vx) + 0.001\cdot q_{\text{safe}}(\vx)$}\\
\cmidrule(lr){2-3} \cmidrule(lr){4-5}
\multicolumn{1}{c}{target $p$} & \multicolumn{1}{c}{$\epsilon=0.01$} & \multicolumn{1}{c}{$\epsilon=0.05$} & \multicolumn{1}{c}{$\epsilon=0.01$} & \multicolumn{1}{c}{$\epsilon=0.05$}\\
{\includegraphics[width=0.19\textwidth]{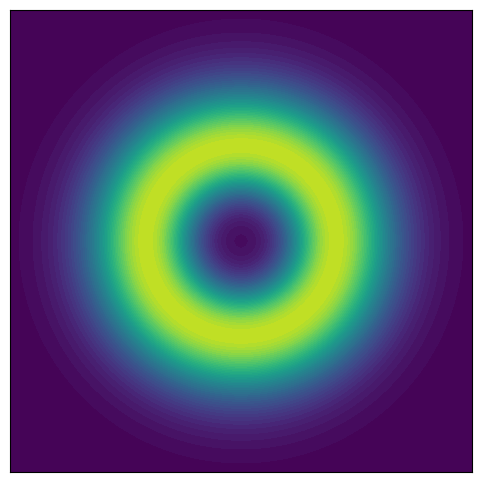}\label{fig:target1}} &
{\includegraphics[width=0.19\textwidth]{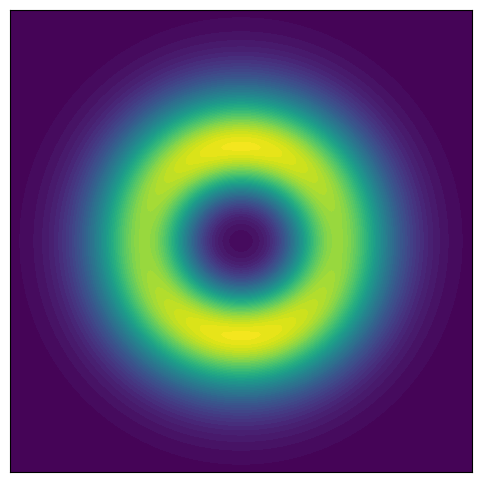}\label{fig:t1p1}} &
{\includegraphics[width=0.19\textwidth]{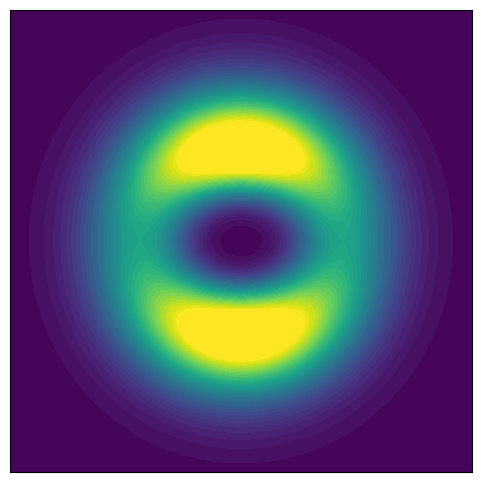}\label{fig:t1p2}} &
{\includegraphics[width=0.19\textwidth]{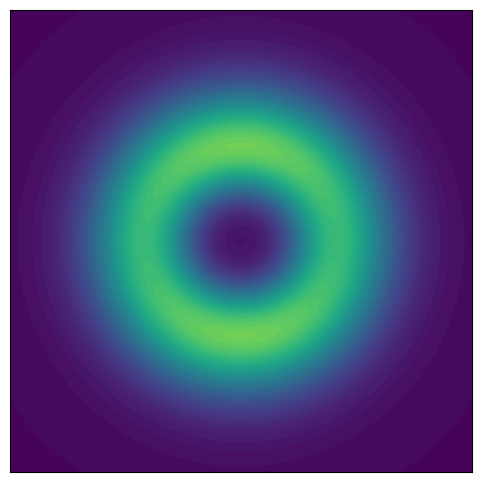}\label{fig:t1p3}} &
{\includegraphics[width=0.19\textwidth]{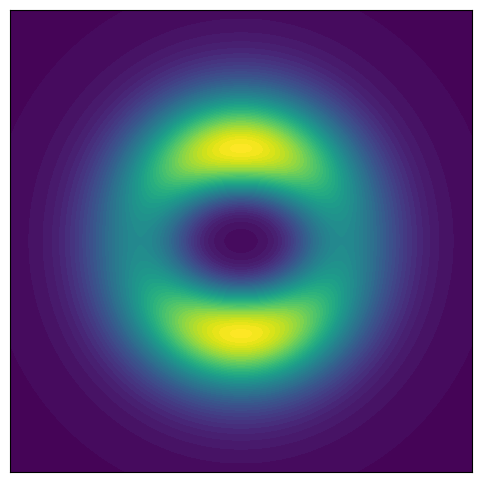}\label{fig:t1p4}} \\
\small$ \widehat{\mathrm{KL}}(p||q) $  & 1.10e-03   & 3.11e-02  &   1.72e-01   & 1.97e-01\\ 
 \small$I^{-1}\sqrt{\widehat{\mathbb{V}}[\widehat{I}]}$ ($\widehat{\text{CoV}}$) & 4.30e-03  & 4.88e-02     &   3.25e-02    &     3.99e-02\\ 
\small$ \log(|\widehat{I} - I|)$ & -1.17e+01 & -9.33e+00  & -9.76e+00 & -9.67e+00\\
\midrule
{\includegraphics[width=0.19\textwidth]{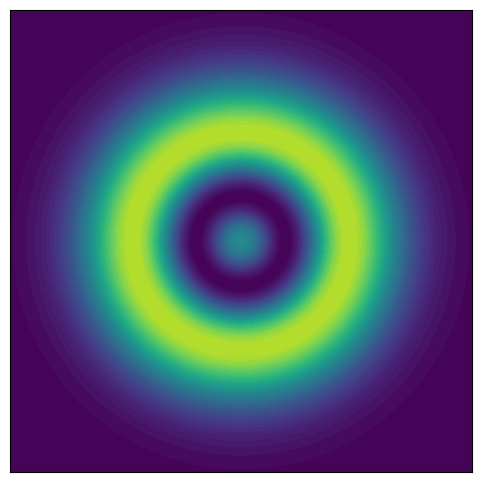}\label{fig:target2}} &
{\includegraphics[width=0.19\textwidth]{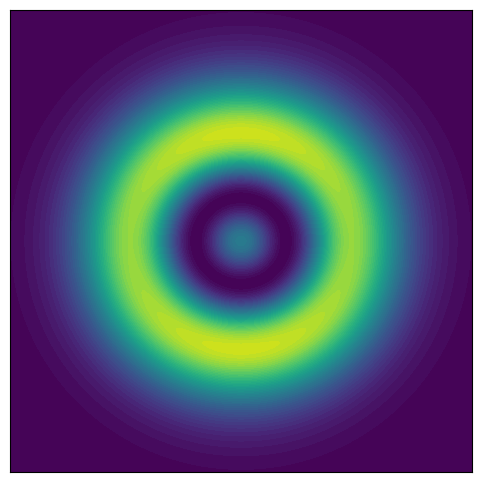}\label{fig:t2p1}} &
{\includegraphics[width=0.19\textwidth]{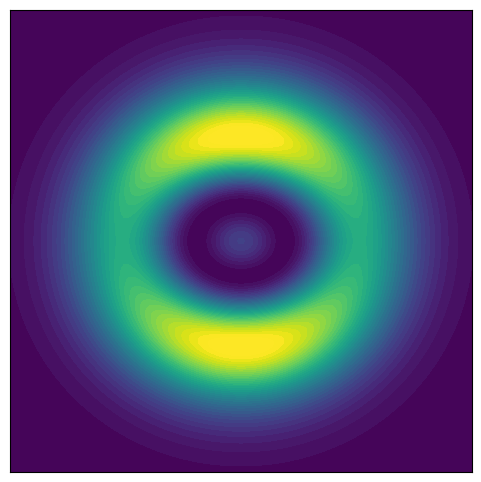}\label{fig:t2p2}} &
{\includegraphics[width=0.19\textwidth]{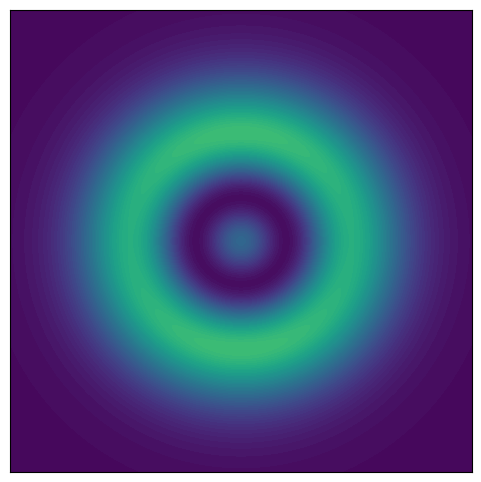}\label{fig:t2p3}} &
{\includegraphics[width=0.19\textwidth]{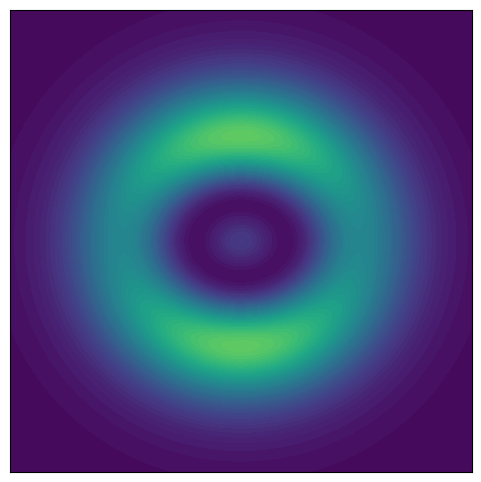}\label{fig:t2p4}}\\
\small$ \widehat{\mathrm{KL}}(p||q) $  & 1.29e-03 & 3.10e-02  &  2.29e-01  & 2.73e-01\\ 
 \small$I^{-1}\sqrt{\widehat{\mathbb{V}}[\widehat{I}]}$ ($\widehat{\text{CoV}}$) & 5.81e+03 & 9.04e+06 & 5.00e-02 & 7.28e-02\\ 
\small$ \log(|\widehat{I} - I|)$ &  -7.39e-01 & 2.84e+00 & -9.51e+00 & -9.22e+00\\
\bottomrule
\end{tabular}
\caption{\textbf{$\Delta\text{Ex}$ can result in high variance when $q$ has low-density valleys, but we can fix that with a ``safe'' component}. We show here that \textbf{(i)} choosing a good proposal for $\method$ does not simply amount to having a small KL divergence, as coefficient of variation (CoV) can be very large even when KL is small and \textbf{(ii)} low-density valleys in $q$ can cause high CoV, but we can fix that with the inclusion of a ``safe'' component. The integral of interest here is the normalizing constant of $p$, i.e., $I=\int \widetilde{p}(\vx)d \vx$. Estimates are computed from $S=15000$ samples, $\text{CoV}$ and average error are based on $100$ estimates, and the KL is estimated from $200000$ samples.
} 
\label{fig:synthetic_proposals}
\end{center}
\end{figure}

\cref{fig:synthetic_proposals} shows two simple, two-dimensional targets and corresponding proposals. We compute estimates based on $S=15000$ samples from the proposal. We repeat the estimation $100$ times and report the estimated coefficient of variation (CoV), given as  $I^{-1}\cdot\sqrt{\widehat{\mathbb{V}}\big[\widehat{I}_{\Delta\text{ExS}}\big]}$, as well as the average estimation error, measured as $\log(|\widehat{I}_{\Delta\text{ExS}}-I|)$.
We further report the the estimated KL divergence between the target and the proposal based on $200000$ samples from $p$ obtained by ARITS.  

As can be seen from the results for the first target, $\Delta\text{ExS}$ can exhibit a low estimated CoV even without a safe component when the proposal does not have a prominent low-density valley. For the second target, $\Delta\text{ExS}$ does result in high variance initially but the inclusion of a safe component drastically improves the estimator's variance and average estimation quality, even with a small mixing proportion $\alpha$. Overall, it is an interesting observation that low KL divergence between the target and the proposal does not necessarily correspond to a low-variance estimator. This raises the question how to find good proposals for $\Delta\text{Ex}$ estimators systematically.

\section{Discussion \& Conclusions}
\label{sec:conclusions}

In this  work, we connected optimal IS proposals with SMMs and examined an unbiased IS estimator, which we term $\method$, that allows us to use SMMs as proposals while avoiding costly auto-regressive inverse transform sampling.
We further derived a variance expression that highlights qualitative differences of $\Delta\text{Ex}$ with standard IS estimators, opening many new possibilities to do adaptive IS \citep{bugallo2017adaptive} with SMMs. 
$\method$ shows promising computational improvements over ARITS in our experiments. 
However, the scalability does not come for free. 
Our estimator can return negative values for a positive quantity, and the UIS estimator can sometimes have high variance when the proposal has low-density valleys. Yet, as a first step towards designing appropriate proposals for $\method$, we found that mixing the SMM with a ``safe'' component can address this problem. 

In upcoming work, we will study adaptive IS strategies tailored to the proposed method. 
For instance, it is natural in the setting of latent variable models to consider a version of the evidence lower bound that exploits the difference representation of an SMM, 
\begin{equation}
    \tag{$\Delta$ELBO}
    \mathbb{E}_{q} \left [ \log \left ( \frac{p(\boldsymbol{x} , \boldsymbol{z} )}{q(\boldsymbol{z})} \right ) \right ] = \frac{Z_{+}}{Z_q} \cdot \mathbb{E}_{q_{+}} \left [ \log \frac{p(\boldsymbol{x} , \boldsymbol{z} )}{q(\boldsymbol{z})}  \right ] - \frac{Z_{-}}{Z_q} \cdot \mathbb{E}_{q_{-}} \left [ \log \frac{p(\boldsymbol{x} , \boldsymbol{z} )}{q(\boldsymbol{z})}  \right ]
\end{equation}
and its several variants \citep{pmlr-v235-hotti24a} when using determistic mixture sampling.
Furthermore, we are investigating an extension of $\method$ to hierarchical mixture models that can be represented in the language of deep circuits \citep{vergari2019tractable,mari2023unifying}.
Lastly, we will perform more experiments to evaluate $\method$ against several baselines and other approximate inference approaches. 

\section*{Acknowledgements}
LZ and AV were supported by the ``UNREAL: Unified Reasoning Layer for Trustworthy ML'' project (EP/Y023838/1) selected by the ERC and funded by UKRI EPSRC. The work of VE is supported by ARL/ARO under grant W911NF-22-1-0235. The authors are grateful to Lorenzo Loconte for useful discussion about the implementation of ARITS and the cirkit library, and to Adrián Javaloy for early helpful discussions around SMMs.

\newpage
\section*{Contributions}
NB and AV initially discussed a preliminary idea of using circuits for IS. NB suggested to use the difference representation of SMMs for IS and together with LZ then formalized  $\method$ which was later discussed with AV and VE. LZ highlighted the instability of  $\method$, proposed the inclusion of a safe component to fix it, and suggested the stratified variant of $\method$. NB and LZ jointly provided the proofs and derivations and designed experiments with the help of AV. LZ was responsible for implementing and carrying out experiments with help from NB. NB led the writing with LZ and help from AV and VE. AV supervised and provided feedback for all the steps of the project.

\clearpage 
\bibliographystyle{iclr2025_conference}
\bibliography{refs}

\newpage
\appendix
\section{Proofs}\label{app:proofs}
\subsection{Variance derivation}\label{app:variance}
Under the assumptions in \cref{proposition:properties}, we have
\begin{align}
    \mathbb{V}_{\substack{\vx_+ \sim q_+\\\vx_- \sim q_-}}[\widehat{I}_{\method}] &= \mathbb{V}_{q_{+}}\left[ \frac{Z_+}{Z_q}\frac{1}{S_{+}} \sum\nolimits_{s=1}^{S_{+}}  f(\vx_{+}^{(s)}) w(\vx_{+}^{(s)}) \right ] + \mathbb{V}_{q_{-}}\left [\frac{Z_-}{Z_q}\frac{1}{S_{-}} \sum\nolimits_{s=1}^{S_{-}} f(\vx_{-}^{(s)})w(\vx_{-}^{(s)}) \right ] \\
    &= \frac{Z_+^2}{Z_q^2} \frac{1}{S_{+}} \left ( \mathbb{E}_{q_{+}}\left [ (f(\vx_{+}) w(\vx_{+}))^2 \right ] -  \left (\mathbb{E}_{q_{+}}\left [ f(\vx_{+}) w(\vx_{+}) \right ] \right )^2 \right ) + \\
    & \frac{Z_-^2}{Z_q^2} \frac{1}{S_{-}} \left ( \mathbb{E}_{q_{-}}\left [ (f(\vx_{-}) w(\vx_{-}))^2 \right ] -  \left (\mathbb{E}_{q_{-}}\left [ f(\vx_{-}) w(\vx_{-}) \right ] \right )^2 \right ) ,
\end{align}

where we used the independence of $\vx_{+}$ and $\vx_{-}$ as well as the fact that both $\{ \vx_{+}^{(s)}\} \sim q_{+}$ and $\{ \vx_{-}^{(s)}\} \sim q_{-}$ are sampled i.i.d. When $f(\vx) = 1$ and we replace $p(\vx)$ with its unnormalized version $\widetilde{p}(\vx)$, we obtain the expression in \cref{proposition:properties}. 

\subsection{Unbiasedness and consistency of $\Delta\text{Ex}$ for UIS}\label{app:unbiasedness}
We show that our constructed $\Delta\text{Ex}$ estimator is unbiased for UIS as stated in \cref{proposition:properties}. Note that the standard MC setting, as benchmarked in Table \ref{tab:exp_1}, is the special case of $p=q$.

Recall that our estimator is given as
\begin{equation*}\label{eq:our_estimator}
\widehat{I}_{\method} = \frac{Z_{+}}{Z_q}\frac{1}{S_{+}} \sum\nolimits_{s=1}^{S_{+}}  f(\vx_{+}^{(s)}) w(\vx_{+}^{(s)}) - \frac{Z_{-}}{Z_q}\frac{1}{S_{-}} \sum\nolimits_{s=1}^{S_{-}} f(\vx_{-}^{(s)})w(\vx_{-}^{(s)}), \ \text{where}\ 
\begin{array}{l}
\vx_{+}^{(s)}\sim q_{+}(\vx_{+}) \\[2pt]
\vx_{-}^{(s)} \sim q_{-}(\vx_{-})
\end{array}.
\end{equation*}
Note that $q(\boldsymbol{x})=Z_q^{-1}\Big(Z_+q_+(\boldsymbol{x}) - Z_-q_-(\boldsymbol{x})\Big)$ and $\{ \vx_{+}^{(s)}\} \sim q_{+}$ and $\{ \vx_{-}^{(s)}\} \sim q_{-}$ are sampled i.i.d. 
Assuming that $\int |f(\vx)| p(\vx) dx < \infty$ and $q(\vx) \neq 0$ almost-everywhere in the support of $q_{+}$ and $q_{-}$, we have
\begin{align}
\mathbb{E}_{\substack{\vx_+ \sim q_+\\\vx_- \sim q_-}}[\widehat{I}_{\method}] =& \mathbb{E}_{\substack{\vx_+ \sim q_+\\\vx_- \sim q_-}}\left[\frac{Z_+}{Z_q} \frac{1}{S_{+}} \sum_{s=1}^{S_+}  f(\boldsymbol{x}_{+}^{(s)}) \frac{p(\boldsymbol{x}_{+}^{(s)})}{q(\boldsymbol{x}_{+}^{(s)})} - \frac{Z_-}{Z_q}\frac{1}{S_{-}} \sum_{s^{\prime}=1}^{S_-} f(\boldsymbol{x}_{-}^{(s^{\prime})}) \frac{p(\boldsymbol{x}_{-}^{(s^{\prime})})}{q(\boldsymbol{x}_{-}^{(s^{\prime})})}\right]\\
= & \mathbb{E}_{q_+}\Big[\frac{Z_+}{Z_q} \frac{1}{S_{+}} \sum_{s=1}^{S_+}  f(\boldsymbol{x}_{+}^{(s)}) \frac{p(\boldsymbol{x}_{+}^{(s)})}{q(\boldsymbol{x}_{+}^{(s)})}\Big] - \mathbb{E}_{q_-}\Big[\frac{Z_-}{Z_q} \frac{1}{S_{-}} \sum_{s^{\prime}=1}^{S_-} f(\boldsymbol{x}_{-}^{(s^{\prime})}) \frac{p(\boldsymbol{x}_{-}^{(s^{\prime})})}{q(\boldsymbol{x}_{-}^{(s^{\prime})})}\Big]\\
= &\frac{Z_+}{Z_q} \frac{1}{S_{+}}\mathbb{E}_{q_+}\Big[ \sum_{s=1}^{S_+}  f(\boldsymbol{x}_{+}^{(s)}) \frac{p(\boldsymbol{x}_{+}^{(s)})}{q(\boldsymbol{x}_{+}^{(s)})}\Big] - \frac{Z_-}{Z_q} \frac{1}{S_{-}}\mathbb{E}_{q_-}\Big[\sum_{s^{\prime}=1}^{S_-} f(\boldsymbol{x}_{-}^{(s^{\prime})}) \frac{p(\boldsymbol{x}_{-}^{(s^{\prime})})}{q(\boldsymbol{x}_{-}^{(s^{\prime})})}\Big]\\
= & \frac{Z_+}{Z_q} \frac{1}{S_{+}}\sum_{s=1}^{S_+} \mathbb{E}_{q_+}\Big[f(\boldsymbol{x}_{+}^{(s)}) \frac{p(\boldsymbol{x}_{+}^{(s)})}{q(\boldsymbol{x}_{+}^{(s)})}\Big] - \frac{Z_-}{Z_q} \frac{1}{S_{-}}\sum_{s^{\prime}=1}^{S_-} \mathbb{E}_{q_-}\Big[f(\boldsymbol{x}_{-}^{(s^{\prime})}) \frac{p(\boldsymbol{x}_{-}^{(s^{\prime})})}{q(\boldsymbol{x}_{-}^{(s^{\prime})})}\Big]\\
= & \frac{Z_+}{Z_q}\mathbb{E}_{q_+}\Big[f(\boldsymbol{x}_{+}) \frac{p(\boldsymbol{x}_{+})}{q(\boldsymbol{x}_{+})}\Big] - \frac{Z_-}{Z_q} \mathbb{E}_{q_-}\Big[f(\boldsymbol{x}_{-}) \frac{p(\boldsymbol{x}_{-})}{q(\boldsymbol{x}_{-})}\Big]\\
= & \frac{Z_+}{Z_q}\int f(\boldsymbol{x}) \frac{p(\boldsymbol{x})}{q(\boldsymbol{x})}q_+(\vx)d\boldsymbol{x} - \frac{Z_-}{Z_q} \int f(\boldsymbol{x}) \frac{p(\boldsymbol{x})}{q(\boldsymbol{x})}q_-(\boldsymbol{x})d\boldsymbol{x} \\
= &\int f(\boldsymbol{x}) \frac{p(\boldsymbol{x})}{q(\boldsymbol{x})}\frac{1}{Z_q}\Big(Z_+q_+(\boldsymbol{x}) - Z_-q_-(\boldsymbol{x})\Big)d\boldsymbol{x} \\
= &\int f(\boldsymbol{x}) \frac{p(\boldsymbol{x})}{q(\boldsymbol{x})}q(\boldsymbol{x})d\boldsymbol{x} \\
=& \int f(\boldsymbol{x}) p(\boldsymbol{x})d\boldsymbol{x} = I. 
\end{align}
Under the same assumptions as above and by directly applying the strong law of large numbers, we have almost sure (a.s.) consistency, as
\begin{equation}
    \mathbb{P}_{\substack{\vx_+ \sim q_+\\\vx_- \sim q_-}}\left ( \lim_{\substack{S_+ \rightarrow +\infty\\S_- \rightarrow +\infty}} \widehat{I}_{\method} = I \right ) = 1 .
\end{equation}

\subsection{Variance Reduction by Stratified Sampling}\label{app:sampling_techniques}
Given an overall sampling budget $S$ and a (normalized) additive mixture $q(\vx) = \sum_{k=1}^K \alpha_k q_k(\vx)$, with $\alpha_k \geq 0$ and $\sum_{k=1}^K \alpha_k = 1$,  
\textbf{\textit{ancestral sampling}} first selects a component for each of the $S$ samples by drawing from a categorical distribution, i.e. $C^{(s)} \sim \text{Cat}(\alpha_1, ..., \alpha_k)$ for $s \in \{1, ..., S\}$, and then samples from the resulting component, $\vx^{(s)} \sim q_{C^{(s)}}$.

On the other hand,
\textbf{\textit{stratified sampling}} deterministically assigns the number of samples drawn from each component in proportion to its assigned mixture weight. The $k$-th component receives a sampling budget of $S_k \approx  \alpha_k S$. 

Since our estimator is a difference of two independent estimators,
\begin{equation*}
\widehat{I}_{\method} = \frac{Z_{+}}{Z_q}\frac{1}{S_{+}} \sum\nolimits_{s=1}^{S_{+}}  f(\vx_{+}^{(s)}) w(\vx_{+}^{(s)}) - \frac{Z_{-}}{Z_q}\frac{1}{S_{-}} \sum\nolimits_{s=1}^{S_{-}} f(\vx_{-}^{(s)})w(\vx_{-}^{(s)}), \ \text{where}\ 
\begin{array}{l}
\vx_{+}^{(s)}\sim q_{+}(\vx_{+}) \\[2pt]
\vx_{-}^{(s)} \sim q_{-}(\vx_{-}) \nonumber 
\end{array} , 
\end{equation*}
the variance is a sum of two variances 
\begin{align*}
    \mathbb{V}_{\substack{\vx_+ \sim q_+\\\vx_- \sim q_-}}[\widehat{I}_{\method}] = \frac{Z_{+}^2}{Z_{q}^2}\mathbb{V}_{q_{+}}\left [\frac{1}{S_{+}} \sum_{s=1}^{S_+}  f(\boldsymbol{x}_{+}^{(s)}) \frac{p(\boldsymbol{x}_{+}^{(s)})}{q(\boldsymbol{x}_{+}^{(s)})} \right ] + \frac{Z_{-}^2}{Z_{q}^2}\mathbb{V}_{q_{-}}\left [\frac{1}{S_{-}} \sum_{s=1}^{S_{-}}  f(\boldsymbol{x}_{-}^{(s)}) \frac{p(\boldsymbol{x}_{-}^{(s)})}{q(\boldsymbol{x}_{-}^{(s)})} \right ]  .
\end{align*}
Since both $q_{+}$ and $q_{-}$ are additive MMs, we can apply the known result that the variance under stratified sampling cannot be greater than that of ancestral sampling for both terms in the above (see, e.g., \citet{elvira2019generalized} comparing the $\mathrm{S}_1$ and $\mathrm{R}_1$ schemes). As a result, the variance for $\Delta\text{ExS}$ cannot be greater than that for $\Delta\text{ExA}$. 

\section{Experimental Setup}\label{sec:experiments_appendix}

We implement all of our experiments in Python using the \texttt{cirkit} library \citep{The_APRIL_Lab_cirkit_2024}. 
Experiments reporting runtime are run on a single NVIDIA L40 (45GiB VRAM) GPU each and runtimes are measured using \texttt{time.perf\_counter}.

\subsection{Experiment 1: Comparison of $\Delta\text{ExS}$ and ARITS}
In our first set of experiments, depicted in \cref{tab:exp_1}, our sampling distribution is a squared SMM with Gaussian inputs. The means are initialized with a standard normal distribution and standard deviations are drawn from a $\text{Unif}(2, 3)$ distribution. The mixture weights are initialized with a $\text{Unif}(-1, 1)$ distribution. We repeat the random initialization until the generated model has at least one negatively weighted component after squaring. 

The target function $f$ is initialized with 100 Gaussian components for all settings. The means are initialized using a standard normal distribution and the standard deviations are sampled from a $\text{Unif}(1, 2)$ distribution. The weights of the sum layer are sampled from a $\text{Unif}(10000, 100000)$ distribution to encourage a non-zero target expectation in high dimensions.

\subsection{Experiment 2: $\Delta\text{ExS}$ for UIS}\label{sec:exp_2_init_appendix}

In this experiment, we try to predict the normalizing constants of two target densities given by squared SMMs and denoted by $p_1$ and $p_2$. The targets $p_1$ and $p_2$ are constructed by squaring and re-normalizing an (unnormalized) SMM with the components $\mathcal{N}([0,0]^T, 0.6^2 \cdot \mathbf{1})$ and $\mathcal{N}([0,0]^T,\mathbf{1})$, where $\mathbf{1}$ is the two-dimensional identity, and mixture weights $\alpha_1=0.12$, $\alpha_2=-0.36$ (for $p_1$) and $\alpha_1=0.16$, $\alpha_2=-0.36$ (for $p_2$) respectively. Since in this experiment we are estimating the normalizing constant of the target distribution, $f(\vx) = 1$.

\subsection{ARITS Sampling}\label{sec:appendix_ARITS}
In our implementation of ARITS, we use a binary search for numerically inverting the conditional CDF. We choose the start and end points of the binary search as -100 and 100 respectively across all settings\footnote{To ensure the validity of the algorithm, we always check whether these bounds result in a (conditional) CDF of 0 and 1 respectively.}. The search is stopped once the upper and lower bound for each sample differ by no more than $10^{-6}$. Pseudocode for our ARITS implementation is given in Algorithm \ref{alg:ARITS}. Note that computing the conditional CDF in the algorithm is tractable \citep{vergari2021compositional}.

The runtime and performance reported for ARITS in Table \ref{tab:exp_1} could likely be improved via a more efficient implementation of the algorithm, parallelization, or a different choice of hyperparameters (i.e., start and end point for binary search, stopping criterion, sampling batch size). Nevertheless, sampling with ARITS is inherently sequential in the number of features.

Further, note that while continuing the binary search until all samples in the batch reached the specified tolerance, as in our implementation, might seem inefficient, we found that generally all samples reached the desired tolerance in the same iteration.

\begin{algorithm}[ht!]
    \SetKwInOut{Input}{Input}
    \SetKwInOut{Output}{Output}
    \Input{(normalized) sampling distribution $q_{\text{SMM}}$,\newline 
    number of samples $S$, \newline number of features $D$,\newline
    initial upper bound $B$, \newline
    initial search lower bound $L$,\newline
    tolerance for stopping binary search $T$
    }
    \Output{set of $S$ samples from $p_{\text{SMM}}$ $\boldsymbol{X}_S$}
    $\boldsymbol{X}_S = ()$\; 
    \For{$i \in \{1, ..., d\}$}{
        $\boldsymbol{u} \gets (u_1, ..., u_S) \sim \text{Unif}(0,1)^S$\;
        $\boldsymbol{L} \gets (L, ..., L), \; |\boldsymbol{L}| = S$\;
        $\boldsymbol{B} \gets (B, ..., B), \; |\boldsymbol{B}| = S$\;
        \If{$i > 1$}{
        \tcc{Pre-compute evidence of observations made so far}
        $\boldsymbol{e} \gets q_{\text{SMM}}(\boldsymbol{x}_1, ..., \boldsymbol{x}_{i-1})$\;
        }
        \Else{
            $\boldsymbol{e} \gets (1, ..., 1), \; |\boldsymbol{e}| = S$
        }
        \tcc{Perform binary search to numerically invert CDF}
        \While{Any $|\boldsymbol{L} - \boldsymbol{B}| > T$}{
            $\boldsymbol{M} \gets \boldsymbol{L} + (\boldsymbol{B}-\boldsymbol{L})/2$\;
            \tcc{Compute conditional CDF at midpoint}
            $\boldsymbol{c} \gets q_{\text{SMM}}(\boldsymbol{x}_i \leq \boldsymbol{M}, \; \boldsymbol{x}_1, ..., \boldsymbol{x}_{i-1})/\boldsymbol{e}$
            
            \tcc{Update upper and lower bounds for $\boldsymbol{x}_i$}
            \For{$s \in \{1,...,S\}$}{
                \If{$c_s > u_s$}{
                   $B_s \gets M_s$
                }
                \Else{
                    $L_s \gets M_s$
                }
            }
            
        }
        \tcc{Re-compute midpoint and append i-th dimension to sample}
        $\boldsymbol{M} \gets \boldsymbol{L} + (\boldsymbol{B}-\boldsymbol{L})/2$\;
        $\boldsymbol{X}_S \gets (\boldsymbol{X}_S, \boldsymbol{M})$       
    }
    \Return{$\boldsymbol{X}_S$}
    \caption{\textbf{Pseudocode for the ARITS implementation used in our experiments.} \newline For Table \ref{tab:exp_1}, $L=-100$, $B=100$, $T=10^{-6}$.}\label{alg:ARITS}
\end{algorithm}

\newpage 

\section{Additional Monte Carlo Experiments}\label{sec:appendix_additional_exps}

\subsection{Empirical Variance Comparison of $\Delta\text{ExS}$ and $\Delta\text{ExA}$}\label{app:variance_comparison}
As discussed in \cref{subsec:estimator} and \cref{app:sampling_techniques}, we expect $\Delta\text{ExS}$, which uses stratified mixture sampling, to have smaller or equal variance than $\Delta\text{ExA}$, which relies on ancestral sampling.
\cref{tab:sample_allocation} gives an empirical comparison of the variance of the two estimators for the same setup as in \cref{sec:runtime_comparison} and \cref{tab:exp_1}. Estimators of the variance are computed from 100 $\Delta\text{Ex}$ estimators each and the table reports the variance ratio of $\Delta\text{ExS}$ and $\Delta\text{ExA}$, i.e. $\widehat{\mathbb{V}}[\Delta\text{ExS}] \cdot (\widehat{\mathbb{V}}[\Delta\text{ExA}])^{-1}$. A ratio lower than 1 hence indicates a variance reduction by stratified sampling.

\begin{table}
\begin{tabular}{lllll}
\toprule
$\boldsymbol{d}$ & $\boldsymbol{S}$ & $\widehat{\mathbb{V}}[\Delta\text{ExS}] \cdot (\widehat{\mathbb{V}}[\Delta\text{ExA}])^{-1}$ & 
$\widehat{\mathbb{V}}[\Delta\text{ExS}] \cdot (\widehat{\mathbb{V}}[\Delta\text{ExA}])^{-1}$
& $\widehat{\mathbb{V}}[\Delta\text{ExS}] \cdot (\widehat{\mathbb{V}}[\Delta\text{ExA}])^{-1}$\\ \midrule
16 & 10000 & 1.133 $\pm$ 0.346 & 1.033 $\pm$ 0.183 & 1.000 $\pm$ 0.000 \\
32 & 10000 & 1.133 $\pm$ 1.196 & 1.333 $\pm$ 1.322 & 0.933 $\pm$ 0.868 \\
64 & 10000 & 49.267 $\pm$ 200.313 & 8.733 $\pm$ 17.875 & 75.467 $\pm$ 191.528 \\\bottomrule
\end{tabular}
\caption{\textbf{Stratified sampling does not provide noticeable variance reduction compared to ancestral sampling in our experiments.} Results for MC comparing $\Delta\text{ExS}$ (stratified sampling) with $\Delta\text{ExA}$ (ancestral sampling), for a varying number for features ($d$) and components ($K$). The variance estimators are computed based on 100 $\Delta\text{Ex}$ estimators. Results are averaged over 30 initializations of $p$ and $f$ (mean $\pm$ stddev). The experimental setup is as in \cref{tab:exp_1}.}\label{tab:variance_comparison}
\end{table}

Interestingly, we do not observe any variance reduction by using stratified sampling empirically. 
The difference between the two techniques could be potentially more apparent in a different experimental setup. Note that for $d=64$, the estimators of the variance are likely unreliable due to small values of the estimators. For instance, ground-truth values for $d=64$, $K=2$ can be in the order of $10^{-47}$. 

\subsection{Effect of Sampling Budget Allocation on $\Delta\text{Ex}$}\label{app:sampling_budget}

In \cref{tab:sample_allocation} we compare our proposed sample allocation for $q_+$ and $q_-$, i.e. $S_+ = \big\lfloor\frac{Z_+}{Z_+ + Z_-} S\big\rfloor$ and $S_- = \big\lfloor\frac{Z_-}{Z_+ + Z_-} S\big\rfloor$ (see \cref{subsec:estimator}), with equal sampling splitting, i.e. $S_+=\big\lfloor\frac{S}{2}\big\rfloor$ and $S_-=\big\lfloor\frac{S}{2}\big\rfloor$. We denote the estimator using equal sample splitting as $\Delta\text{ExS} \; (\text{Eq.})$. We use the same experimental setup as in \cref{sec:runtime_comparison}, hence the results for $\Delta\text{ExS}$ coincide with \cref{tab:exp_1}. We find that the two sampling schemes achieve similar mean errors for this set of examples.

\begin{table}[htp!]
\resizebox{\textwidth}{!}{
\begin{tabular}{lrrrrr}
\toprule
& & &\multicolumn{3}{c}{\textbf{Number of components} ($\boldsymbol{K}$)}\\
\cmidrule(lr){4-6}
& & & $\boldsymbol{2}$ & $\boldsymbol{4}$ & $\boldsymbol{6}$\\\cmidrule(lr){4-4} \cmidrule(lr){5-5} \cmidrule(lr){6-6}
\textbf{Method} & $\boldsymbol{d}$ & $\boldsymbol{S}$ & $\log(| \widehat{I} -  I|)$ & $\log(| \widehat{I} -  I|)$ & $\log(| \widehat{I} -  I|)$ \\ \midrule
$\Delta\text{ExS}$ & 16 & 10000 & -18.650 $\pm$ 1.291 & -18.240 $\pm$ 1.314 & -17.942 $\pm$ 1.199 \\
$\Delta\text{ExS (Eq.)}$ & 16 & 10000 & -18.070 $\pm$ 0.900 & -18.300 $\pm$ 1.196 & -18.047 $\pm$ 1.488 \\ \hline
$\Delta\text{ExS}$ & 16 & 100000 & -19.722 $\pm$ 1.041 & -19.231 $\pm$ 1.228 & -19.009 $\pm$ 1.379 \\
$\Delta\text{ExS (Eq.)}$ & 16 & 100000 & -19.325 $\pm$ 1.016 & -19.206 $\pm$ 0.933 & -18.874 $\pm$ 1.219 \\\hline
$\Delta\text{ExS}$ & 16 & 300000 & -19.937 $\pm$ 0.797 & -19.743 $\pm$ 1.070 & -19.491 $\pm$ 0.931 \\
$\Delta\text{ExS (Eq.)}$ & 16 & 300000 & -20.003 $\pm$ 0.920 & -19.790 $\pm$ 1.338 & -19.196 $\pm$ 1.175 \\ \midrule
$\Delta\text{ExS}$ & 32 & 10000 & -48.024 $\pm$ 1.344 & -47.231 $\pm$ 1.059 & -47.273 $\pm$ 1.015 \\
$\Delta\text{ExS (Eq.)}$ & 32 & 10000 & -47.377 $\pm$ 0.964 & -47.138 $\pm$ 1.056 & -46.845 $\pm$ 0.702 \\\hline
$\Delta\text{ExS}$ & 32 & 100000 & -48.395 $\pm$ 1.068 & -48.092 $\pm$ 0.771 & -48.426 $\pm$ 1.734 \\
$\Delta\text{ExS (Eq.)}$ & 32 & 100000 & -47.908 $\pm$ 1.022 & -48.068 $\pm$ 1.094 & -48.501 $\pm$ 1.337 \\\hline
$\Delta\text{ExS}$ & 32 & 300000 & -48.729 $\pm$ 0.835 & -48.671 $\pm$ 1.758 & -48.791 $\pm$ 1.007 \\
$\Delta\text{ExS (Eq.)}$ & 32 & 300000 & -48.250 $\pm$ 0.909 & -48.025 $\pm$ 0.918 & -48.561 $\pm$ 0.833 \\ \midrule
$\Delta\text{ExS}$ & 64 & 10000 & -108.129 $\pm$ 1.280 & -107.338 $\pm$ 0.854 & -107.444 $\pm$ 0.827 \\
$\Delta\text{ExS (Eq.)}$ & 64 & 10000 & -107.973 $\pm$ 0.989 & -107.226 $\pm$ 0.665 & -107.518 $\pm$ 1.371 \\\hline
$\Delta\text{ExS}$ & 64 & 100000 & -108.000 $\pm$ 1.066 & -107.431 $\pm$ 0.692 & -107.672 $\pm$ 0.782 \\
$\Delta\text{ExS (Eq.)}$ & 64 & 100000 & -108.187 $\pm$ 1.281 & -107.386 $\pm$ 0.741 & -107.644 $\pm$ 0.967 \\\hline
$\Delta\text{ExS}$ & 64 & 300000 & -108.354 $\pm$ 1.436 & -107.569 $\pm$ 1.215 & -108.052 $\pm$ 1.372 \\
$\Delta\text{ExS (Eq.)}$ & 64 & 300000 & -108.044 $\pm$ 1.399 & -107.816 $\pm$ 0.938 & -107.858 $\pm$ 0.885 \\
\bottomrule
\end{tabular}}
\caption{\textbf{$\Delta\text{ExS}$ performs similarly with our proposed sample allocation and equal sample splitting.} Results for MC comparing $\Delta\text{ExS}$ with our proposed sample allocation and equal sample splitting $\text{(Eq.)}$, for a varying number for features ($d$) and components ($K$) as well as different sampling budgets ($S$). The error is given as $\log(| \widehat{I} -  I|)$, hence lower is better. Results are averaged over 30 initializations of $p$ and $f$ (mean $\pm$ stddev). The experimental setup is as in \cref{tab:exp_1}.}\label{tab:sample_allocation}
\vspace*{-\baselineskip}
\end{table}

\section{Comparison of ARITS and $\Delta\text{ExS}$ samples}\label{app:sample_comparison}
\begin{figure}[!t]
\begin{center}
\setlength{\subfigcapskip}{0pt}  %
\setlength{\subfigtopskip}{0pt}  %
\setlength{\subfigbottomskip}{0pt}  %
\setlength{\tabcolsep}{2pt}

\begin{tabular}{cccc}
$q_{\text{SMM}}$ & ARITS\\
{\includegraphics[width=0.2\textwidth]{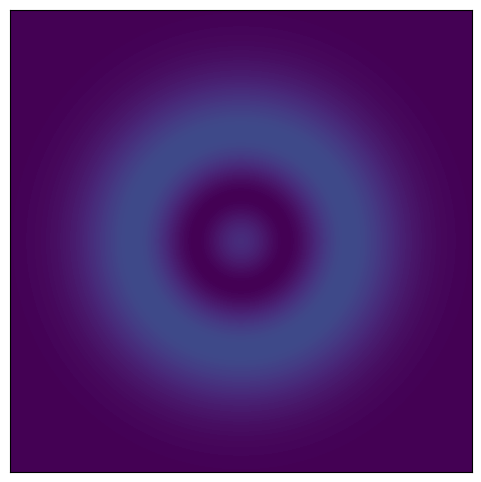}} &
{\includegraphics[width=0.2\textwidth]{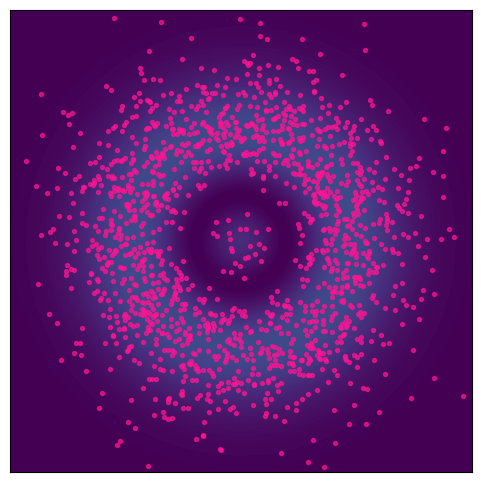}}\\
$q_{+}$ & $\Delta\text{ExS}$ & $q_{-}$ & $\Delta\text{ExS}$\\
{\includegraphics[width=0.2\textwidth]{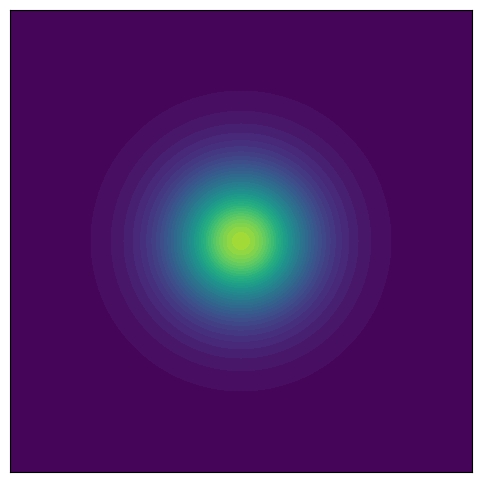}\label{fig:target1}} &
{\includegraphics[width=0.2\textwidth]{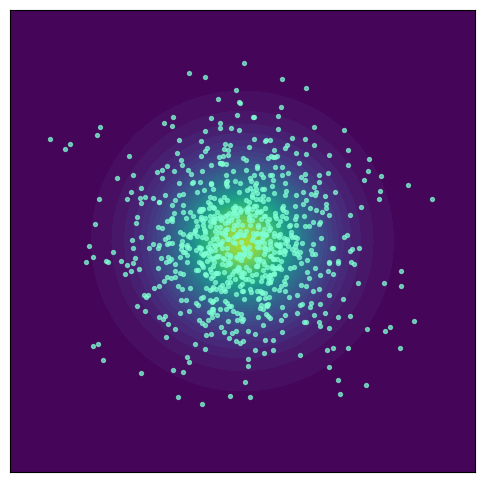}}
&
{\includegraphics[width=0.2\textwidth]{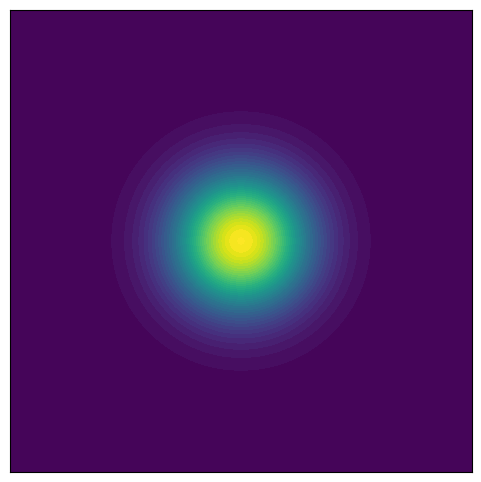}} &
{\includegraphics[width=0.2\textwidth]{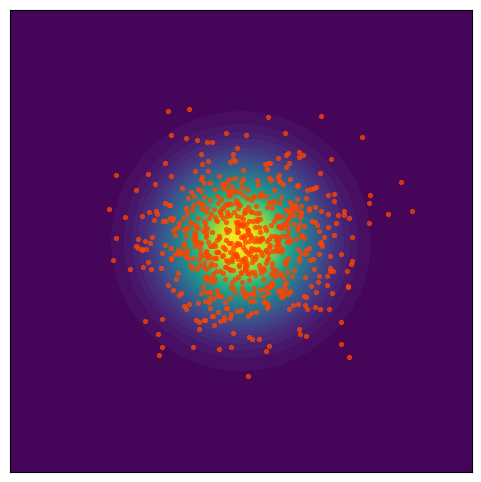}}
\end{tabular}
\caption{\textbf{ARITS samples directly from the full SMM $q_{\text{SMM}}$ while $\Delta\text{ExS}$ samples from $q_+$ and $q_-$ in isolation.} The figure shows 1500 samples obtained by ARITS (first row) and $\Delta\text{ExS}$ (second row) respectively. \emph{All depicted densities are normalized and visualized using the same color map.}
}
\label{fig:sample_comparison}
\end{center}
\end{figure}

Figure \ref{fig:sample_comparison} compares 1500 samples obtained from a squared SMM by ARITS and $\Delta\text{ExS}$. We denote the full SMM by $q_\text{SMM}(\vx) = \frac{1}{Z_q}\big(Z_+ \cdot q_+(\vx) - Z_-\cdot q_-(\vx)\big)$ as in  \cref{eq:diffrep}. ARITS samples directly from $q_{\text{SMM}}$, while $\Delta\text{ExS}$ samples from $q_+$ and $q_-$ in isolation. Note that $\Delta\text{ExS}$ does not sample from $q_\text{SMM}$ but the samples from $q_+$ and $q_-$ can be combined into an unbiased expectation estimator over $q_{\text{SMM}}$ (see \cref{subsec:estimator}). The visualized SMM coincides with the second target in \cref{fig:synthetic_proposals}.

\end{document}